# VERIFICATION OF CONFLICTION AND UNREACHABILITY IN RULE-BASED EXPERT SYSTEMS WITH MODEL CHECKING

Einollah pira[1], Mohammad Reza Zand Miralvand[2] and Fakhteh Soltani[3]

Department of Computer Engineering, Arak University, Arak City, Iran

[1]e-pira@phd.araku.ac.ir
[2]m-miralvand@phd.araku.ac.ir
[3]f-soltani@araku.ac.ir

## ABSTRACT

*It is important to find optimal solutions for structural errors in rule-based expert systems .Solutions to discovering such errors by using model checking techniques have already been proposed, but these solutions have problems such as state space explosion. In this paper, to overcome these problems, we model the rule-based systems as finite state transition systems and express confliction and unreachability as Computation Tree Logic (CTL) logic formula and then use the technique of model checking to detect confliction and unreachability in rule-based systems with the model checker UPPAAL.*

## 1. INTRODUCTION

A rule base is the central part of an expert system that extracts the knowledge from domain experts in the form of inference rules. Structural errors usually appear by augmenting the knowledge base rules. According to [1] , the typical types of structural errors include confliction (conflict rules), unreachability (unreachable rules), subsumption (subsumed rules) , redundancy (redundant rules), and circularity (circular depending rules). But we just focus on the confliction and unreachability in this paper.

Model checking is an automatic method for studying the properties given to a system and their verification [2]. In [3] a solution by using model checking is presented, but it has the following problems:

1) State space explosion: with the increase of rules, the number of states of the model checker increases exponentially, and this makes the model checker is unable to continue his work (out of memory).
2) The model checker has been used in this solution is textual and it makes the importing of rules to the model checker become complicated.

In this paper, to overcome these problems, we model the rule-based systems as finite state transition systems and express confliction and unreachability as Computation Tree Logic (CTL) logic formula and then use the technique of model checking to detect confliction and unreachability in rule-based systems with the model checker UPPAAL.

The rest of the paper is organized as follows. In section 2, related works is presented. In section 3, we briefly introduce the required background.. Section 4 presents our proposed method to detect confliction and unreachability in rule-based systems with the model checker UPPAAL. Finally, we conclude the paper and highlight the future works in section 5.

## 2. RELATED WORKS

Many different techniques have been proposed to detect the structural errors in rule-based systems [4]. Initial works mostly concentrated on the detection of structural errors by checking rules pair-wisely. Recent works focused on detecting structural errors made by implementing multiple rules in longer inference chains. Using some graphical notation such as Petri nets and graphs is a approach in the majority of the recent verification techniques [5]. Some of the mentioned approaches cannot discover structural errors exactly. The approach in [6] could only detect structural errors matching a set of pre-defined syntactic patterns. The approaches in [7,8] did not detect inconsistency errors. The approach in [4] used an adjacency matrix technique, which has a greater computational cost in space and time.

## 3. PRELIMINARIES

In this section, we briefly present the required preliminaries, i.e., Model Checking and UPPAAL.

### 3.1. Model Checking

Model checking is an automatic method for examining the properties given to a system and their verification[9,10-14]. This verification is done by software tools as a model checkers. A model checker thoroughly explores the state space to decide whether the system satisfies the property. The approach is depicted in Figure 1. In a first step, which is called modeling, the system description is converted into the system model. A system description is, for example, a program written in C, Java or Assembly language. A system model is, for example, a Kripke structure, a labeled transition system, or a finite automaton. The requirements have to be manually formalized because they are mostly given in natural language. The result of this formalization is the formal specification given as formulas in a temporal logic such as CTL (Computation Tree Logic).

CTL is a common logic for model checking, that develops propositional logic with specific temporal operators.

The model and the specification are inputs given to the model checker. The model checker uses an exhaustive search over all reachable states of the model to check whether the model satisfies the formula. In the end, it returns a result. The result may be that the model satisfies the formula or that the model does not satisfy the formula together with a counterexample. Due to the state-explosion problem, it may happen that the model checker runs out of memory and does not return a result.

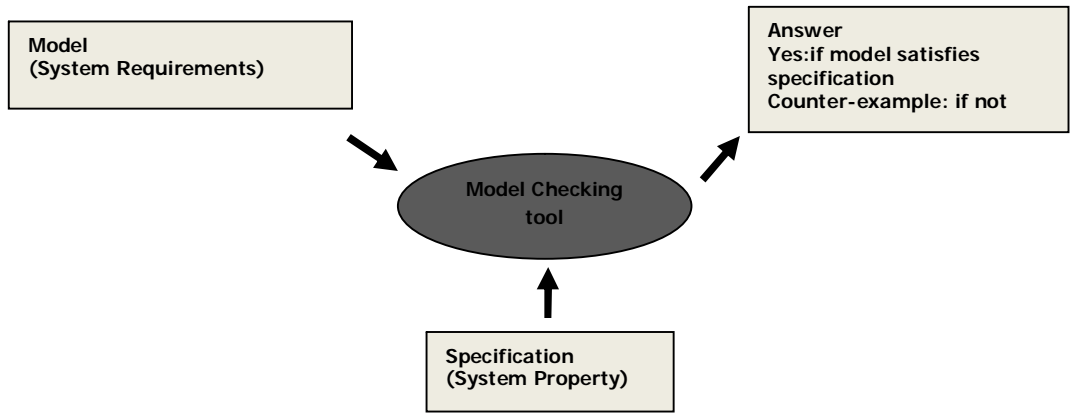

Figure 1: Model checking process [2]

### 3.2. UPPAAL

One of the best tools for the modeling, simulation and verification of real-time systems is UPPAAL [14]. UPPAAL can verify systems that have the following property: they can be modeled as networks of timed automata (TA) expanded with structured data types, integer variables, and channel synchronization. A finite-state machine expanded with clock variables is a TA. UPPAAL expands the definition of TA with extra characteristics. Below are some of these characteristics that are pertinent to our aim [15]:

- **Templates:** A TA is defined as template with optional parameters. Parameters are local variables that are initialized during template instantiation in system declaration.
- **Global variables:** In global declaration section, global variables and user defined functions can be introduced. All templates can access global variables and user defined functions.
- **Expressions**: Three main types of expressions can be existed: (1) Guard expressions, which are evaluated to Boolean and used to limit transitions, they may contain clocks and state variables, (2) Assignment expressions, which are used to set values of clocks and variables, (3) Invariant expressions, which are defined for locations and used to indicate conditions that should be always true in a location.
- **Edges**: Transitions between locations are marked with edges. Each edge specification can consist of four expressions: (1) Select, which assigns a value from a given range to a defined variable, (2) Guard, is a logical expression that if its value is evaluated to true, the corresponding edge is enabled for a location, (3) Synchronization, which describes the synchronization channel and its direction for an edge, and (4) Update, an assignment statements that reset variables and clocks to required values. However, in our paper, we only use two expressions Guard and Update in edges. Figure 2 shows an example: we assume that the system is in a location loc0, if the value of x is 2, then its value will be equal to 4 and the system location will be loc1. Otherwise, its value will not be changed, but the system location will be loc2. Sometimes the edges may not have any expressions.

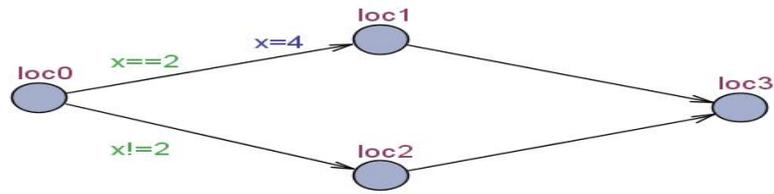

Figure 2: An example in UPPAAL

We use UPPAAL to describe a checking formula that contains a set of properties [16]. The checking formula can be a union of the following (see Figure 3):
- A[] $\varphi$, which means $\varphi$ will invariantly happen
- E<> $\varphi$, which means $\varphi$ will possibly happen
- A<> $\varphi$, which means $\varphi$ will always happen eventually
- E[] $\varphi$, which means $\varphi$ will potentially always happen
- $\varphi$ --> $\psi$, which means $\varphi$ will always lead to $\psi$

Which $\varphi$ and $\psi$ are Boolean expressions defined on locations, integer variables, and clocks constraints.

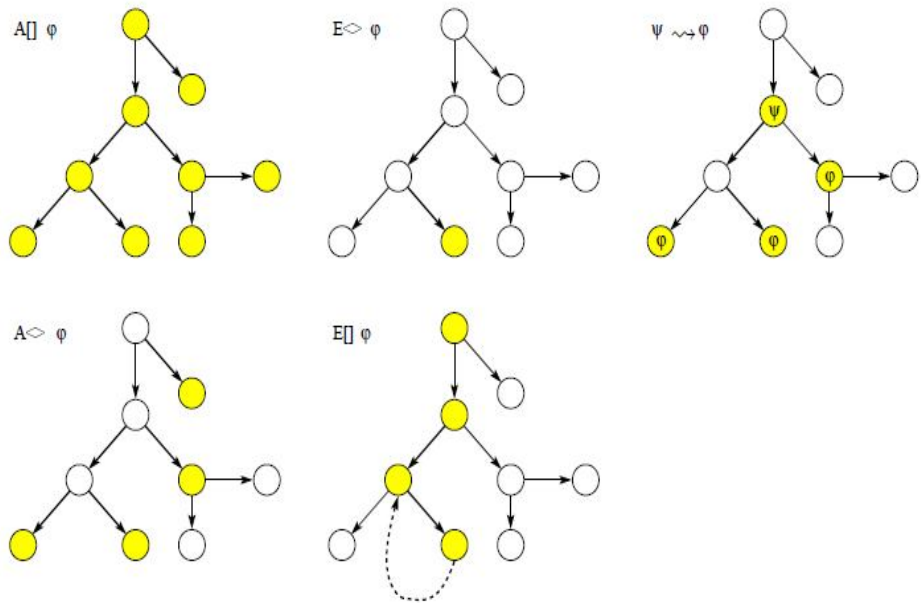

Figure 3: Path formulae supported in Uppaal. The filled states are those for which a given state formulae is true [14].

## 4. OUR PROPOSED METHOD

### 4.1 Explanation of a rule

A rule has the following general form [15]: P → Q , where P and Q are called proposition and deduction respectively. P (or Q) can be an atomic propositional logic formula (a proposition or its negation) or a combined propositional logic formula containing multiple propositions and logical connectives: ∧ and ∨).

For example, a rule base R is defined as follows:

R={
    r0: p0 → p1 ∧ p4
    r1: p1 → ~ p4
    r2: ~p2 → p0 ∧ p1
    r3: p0 ∨ p3 → p4
    r4: p4 → p3
}

### 4.2 Implementation of rules in the UPPAAL

It is assumed that the number of rules in R is m and the number of propositions is n. Each proposition can take three values: 0 (false) , 1 (true) and 2 (nothing). We define an array p with size n to keep the values of propositions, and an array with size m to show that what rules are used. We consider a rule base R as a template. This template consists of the following locations: start (the initial location), rs, rf and ri (i=0..m-1). The corresponding template of the rule base R in the section 4.1 is displayed in Figure 4. When the system goes from location start to location rs, the initp() procedure is called in the local declaration of the template. In this procedure, all entries of array p are set with value 2 (nothing), but the value of p[0] is set 1 because the left-hand side of r0 is p0. However, this procedure is written such that all entries of array p to be initialized only once. For implementing of rule r0: p0 → p1 ∧ p4 , an edge is drawn from location rs to location r0 that its guard expression is p[0]==1 and update expression is p[1]=1 , p[4]=1. Also, an edge is drawn from location r0 to location rf that its update expression is r[0]=true. This edge means that rule r0 is used.

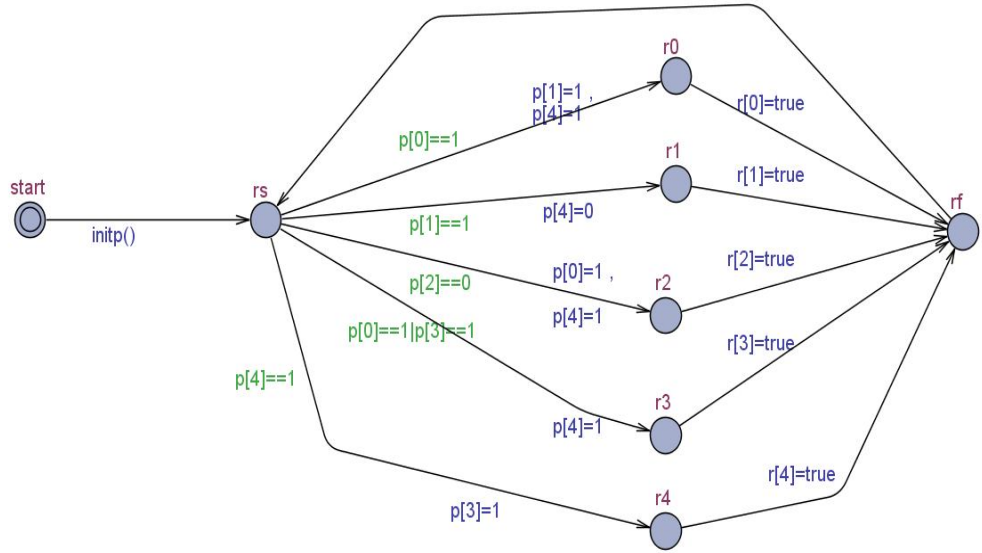

Figure 4: The corresponding template of the rule base R in section 4.1

### 4.3 Verification of confliction

To verify the confliction, it is considered two instances es1 and es2 (also, are called processes ) of the defined template in the section 4.2. Then, we find two rules rx and ry of rule base R that they have the proposition pi and ~pi on their right-hand sides respectively (regardless of the left-hand sides) . For example, in the rule base R of section 4.1, two rules r0 and r1 have the proposition p4 and ~p4 on their right-hand sides respectively. In the verifier section of UPPALL, we insert the following query:
    *E<> es1.r0 and es2.r1*

this means that: eventually, is there the state of the system in which process es1 is in the location r0 and process es2 is in the location r1? If this query is satisfied, two rules r0 and r1 are in conflict with each other, otherwise, two rules mentioned aren't in conflict with each other. In this example, the verifier produces the following response:
    *property is satisfied.*
this means that: two rules r0 and r1 are in conflict with each other.

## 4.4 Verification of unreachability

Similar to the previous section, to verify the unreachability, it is considered two instances es1and es2 (also, are called processes ) of the defined template in the section 4.2. Provided that all rules in the rule base R have been used at least once therefore: r[i]=true ( i=1..m ). So, in the verifier section of UPPALL, we insert the following query (typem is a new type of integer type in the range of 1 to m ):
    *E<> forall (i:typem)  r[i]==true*
this means that: eventually, is there the state of the system in which all r[i] (i:1..m) are true ? If this query is satisfied, all rules in the rule base R have been used at least once, otherwise, some of them are not being used.

In this case, to find out the rule ri is not used, we must check the following query:
    *E<> es1.ri*
this means that: eventually, is there the state of the system in which process es1 is in the location ri? If this query is satisfied, the rule ri has been used at least once, otherwise, this rule has not been used. In this example, the query *E<> es1.r2* has not been satisfied, this means that the rule r2 is unreachable and must be removed from base rule R.

In the end of this section, we want to calculate the total number of system states. Since the defined template in section 4.2 has *3+m* locations (the start, rs and rf locations plus m locations ri's ) and our system have two processes, so the total number of system states for a rule base R with m rules is:
    *N=(3+m)*(3+m)=O($m^2$)*
this means that: the total number of system states is polynomial.

## *5. CONCLUSION AND FUTURE WORKS*

In this paper, we have modeled the rule-based system as finite state transition system and expressed conflict and unreachability as Computation Tree Logic (CTL) logic formula and then used the technique of model checking to detect conflict and unreachability in rule-based systems with the model checker UPPAAL. Our technique has the following advantages:
1) The total number of system states is *O($m^2$)*, so the total number of system states is polynomial.
2) The model checker has been used in this solution is graphical and it makes the importing of rules to the model checker become easy.

An open problem is, we find solutions to detect the other structured errors such as subsumption, redundancy, and circularity.

## BIOGRAPHY

- **Einollah pira**: PhD student in Computer Engineering, Arak University
- **Mohammad Reza Zand Miralvand**: PhD student in Computer Engineering, Arak University
- **Fakhteh Soltani**: Assistant professor in Computer Engineering, Arak University